%% file: 0_ACL_main.tex
\pdfoutput=1
\documentclass[11pt]{article}
\usepackage[]{EMNLP2022}
\usepackage{amsmath}
\usepackage{graphicx}
\usepackage{epstopdf} 
\usepackage{multirow}
\usepackage{times}
\usepackage{latexsym}
\usepackage{amssymb}
\usepackage[T1]{fontenc}
\usepackage[utf8]{inputenc}
\usepackage{hyperref}

\usepackage{microtype}
\usepackage[lined,boxed,commentsnumbered]{algorithm2e}
\usepackage{inconsolata}
\title{A  Hierarchical N-Gram Framework for Zero-Shot Link Prediction}
\author{Mingchen Li\textsuperscript{\normalfont 1}\thanks{\ \  Corresponding author}, Junfan Chen\textsuperscript{\normalfont 2}, Samuel Mensah\textsuperscript{\normalfont 3}\\ \textbf{Nikolaos Aletras}\textsuperscript{3}\textbf{,} \textbf{Xiulong Yang}\textsuperscript{1}\textbf{,} \textbf{Yang Ye}\textsuperscript{1}  \\
        \textsuperscript{1}Georgia State University \textsuperscript{2}Beihang University \textsuperscript{3}University of Sheffield\\ \textsuperscript{1}\{mli33, xyang22, yye10\}@student.gsu.edu, \textsuperscript{2}chenjf@act.buaa.edu.cn\\ \textsuperscript{3}\{s.mensah,n.aletras\}@sheffield.ac.uk
        }
          
\begin{document}
\maketitle

\input{1_abs}

\input{2_introduction}

\input{3_related_work}

\input{4_preliminaries}

\input{5_method}

\input{6_experiment}
\input{7_conclution}

\section*{Acknowledgements}

We would like to thank the anonymous reviewers for their comments and suggestions,
which helped improve the quality of this paper.  We also thank Prof.Richong Zhang from Beihang University for the inspiration of this topic.
SM and NA are supported by a Leverhulme Trust Research Project Grant (No. RPG-2020-148).
% 

\bibliography{anthology}
\bibliographystyle{acl_natbib}
% \subsection{Appendices}
% \appendix \label{appendX}

\clearpage
\section{Appendix}
\label{sec:appendix}
\subsection{ Hierarchical N-gram Graph Building and Node Selection}
\label{Hierarchical N-gram Graph building and node selection}
\begin{table}[ht]
	\centering
	
	\renewcommand\arraystretch{1.3}
	\scalebox{0.7}{
	\begin{tabular} {cc}
		\hline 
		
		\hline	
		Strategy& node order \\ 
		\hline		
	
		Strategy1&(a|p,a,r,t,pa,ar,rt,par,art,part|o,f,of|)\\
		Strategy2&(a,p,a,r,t,o,f | pa,ar,rt,of | par,art| part) \\
		\hline	
	\end{tabular}}
		\caption{Node order of relation "a part of" }
	\label{con:Node number select}
\end{table}

In our work, the n-gram graph from the word level is called the word n-gram graph, and the n-gram graph from the relation level is called the relational n-gram graph (the surface name of the relation contains more than one word).
%%%%%%%%%%
When the surface name of the relation contains more than one word, there are two challenges that need to be solved. The first challenge is how to connect the word n-gram graph to the relation n-grams graph. For the second challenge, when the length of a word is big or the number of words in a relation is big (such as relation "concept: agricultural product growing in state or province" in NELL), the relational n-gram graph will become very big, it is hard for the machine to progress this graph. 

\begin{figure}[t]
% 	\centerin
	\includegraphics[width=0.6\columnwidth]{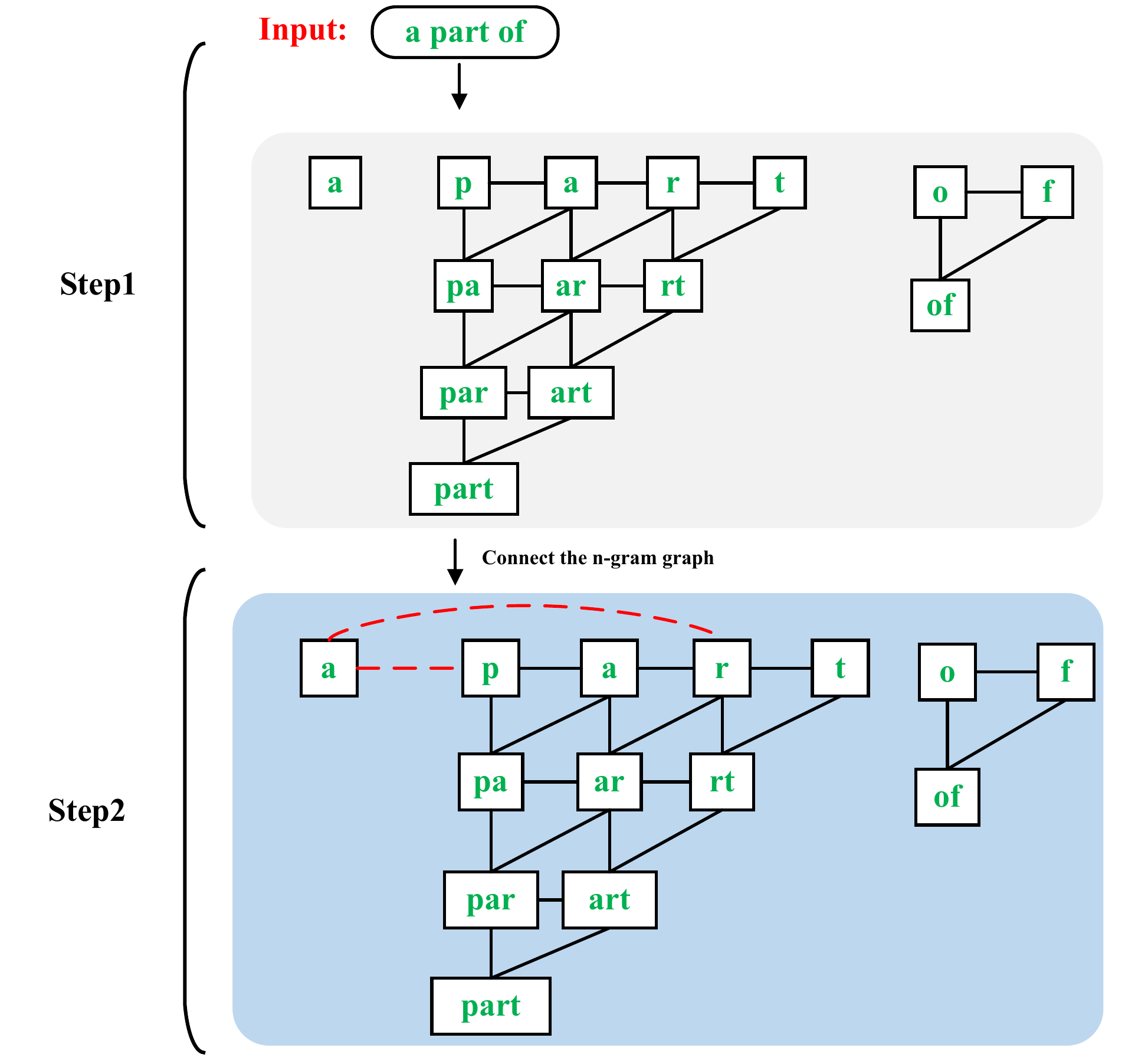} .
	\caption{The building progress of n-gram graph about relation  "a part of"}.
	\label{con:R_gram_graph}
\end{figure}

To solve the first issue,  we connect each word n-gram graph by our proposed two relations.  There are two steps in the building of a relational n-gram graph. Firstly, we split the relation by the space, and build the n-gram for each word. In the second step, we connect each word n-gram graph by  relation \emph{adjoint} and \emph{compositional}. As shown in Figure.\ref{con:R_gram_graph}, for the n-gram graph of first word "a", it appears in the n-gram graph of "part", so it also connects to "p" and "r".

For the second issue, in the relational n-gram graph, we should consider the node order in the first, this way can make sure the word does not lose its internal order information, the node order of the n-gram graph of the word "part" should be "p,a,r,t,pa,ar,...", the order "p,pa,a,ar,r,t,..." is wrong. After that, for considering the run effectiveness and GPU memory, 
we select the fixed number of nodes from left to right based on the order.  Based on the above discussion, we propose two strategies to help order the node. As shown in Table \ref{con:Node number select}.

In Strategy1,
for each word n-gram graph, we rank the node position based on the $n$ in $n$-gram, the position of all $1$-gram nodes all come before the $2$-gram nodes. So, in the first, we list the $1$-gram nodes of relation "a part of",
in the second,  we list the $2$-gram nodes,..
In Strategy2, we list all $n$-grams nodes for each word and then concatenate these nodes together.

\subsection{Out-of-Vocabulary}
\label{out-of-vocalulary}

\begin{table}[ht]
	\centering
	
	\renewcommand\arraystretch{1.3}
	\scalebox{0.7}{
	\begin{tabular} {ccccc}
		\hline 
		
		Strategy& MRR& hits@10&hits@5&hits@1 \\ 
		\hline		
		KGE-word& 0.273& 0.396&0.347&0.202 \\
		HNZSLP&\textbf{0.289} &\textbf{0.413}&\textbf{0.359}&\textbf{0.222}\\
		\hline	

	\end{tabular}}
		\caption{The model performance about the evaluation of out-of-vocabulary problem in  NELL-ZS}
	\label{con:out-of-voc}
\end{table}

By analyzing the dataset of zero-shot link prediction, we found some words in unseen relation are not in the seen relation set, this issue will reduce the performance of tail entity inference. To evaluate the effectiveness of our model, we propose a compared model KGE-word which directly uses the traditional transformer to learn the word information in the surface name of the relation and then uses the KGE model TransE to infer the tail entity. The results are shown in Table \ref{con:out-of-voc}. By comparing with HNZSLP, we can see that using the n-gram graph is better than the way which uses the word information to calculate the relation information. Our n-gram graph can capture the information at different granularities, which is helpful for the knowledge transfer between the seen relation and unseen relation.

\subsection{Node Selection Strategies}
\label{Node Selection Strategies}
\begin{table}[ht]
	\centering
	
	\renewcommand\arraystretch{1.3}
	\scalebox{0.7}{
	\begin{tabular} {ccccc}
		\hline 
		
		\hline	
		Strategy& MRR& hits@10&hits@5&hits@1 \\ 
		\hline		
	 
		strategy1&  0.282& 0.403&0.350&0.216\\
		  strategy2&\textbf{0.289} &\textbf{0.413}&\textbf{0.359}&\textbf{0.222}\\
		\hline	
	\end{tabular}}
		\caption{The performance of HNZSLP in NELL-ZS with different Node order strategies}
	\label{con:Node order}
\end{table}
In this section, we evaluate the performance of different node order strategies. For a fair comparison, we  set $13$-gram, the maximum node number is 90, the epoch of training is 80, and the KGE model is transE. In Table \ref{con:Node order}, the results show that the performance of strategy2  is better than strategy1, which shows that the lower grams are more important than the higher grams.

\end{document}

%% file: 1_abs.tex
\begin{abstract}
% Due to the incompleteness of knowledge graphs (KGs), zero-shot link prediction (ZSLP) which aims to predict unobserved relations in KGs has attracted recent interest from researchers. A common solution is to use textual features of relations (e.g., surface name or textual descriptions) as auxiliary information to bridge the gap between seen and unseen relations. Current approaches learn an embedding for each word token in the text. These methods lack robustness as they suffer from the out-of-vocabulary (OOV) problem. Meanwhile, models built on character n-grams have the capability of generating expressive representations for OOV words.  Thus, in this paper, we propose a \textbf{H}ierarchical \textbf{N}-Gram framework for \textbf{Z}ero-\textbf{S}hot \textbf{L}ink \textbf{P}rediction (HNZSLP), which considers the dependencies among character n-grams of the relation surface name for ZSLP. Our approach works by first constructing a hierarchical n-gram graph on the surface name to model the organizational structure of n-grams that leads to the surface name. A GramTransformer, based on the Transformer is then presented to model the hierarchical n-gram graph to construct the relation embedding for ZSLP. Experimental results show the proposed HNZSLP achieved state-of-the-art performance on two ZSLP datasets. 
% The code is in \href{https://drive.google.com/file/d/1Mro57n-F9P552qW5jPVDQZkdbzOTp26L/view?usp=sharing}{HNZSLP}.

Knowledge graphs  typically contain a large number of entities but often cover only a fraction of all relations between them (i.e., incompleteness). Zero-shot link prediction (ZSLP) is a popular way to tackle the problem by automatically identifying unobserved relations between entities. Most recent approaches use textual features of relations (e.g., surface names or textual descriptions) as auxiliary information to improve the encoded representation. These methods lack robustness as they are bound to support only tokens from a fixed vocabulary and are unable to model out-of-vocabulary (OOV) words. Subword units such as character n-grams have the capability of generating more expressive representations for OOV words. Hence, in this paper, we propose a {\bf H}ierarchical {\bf N}-gram framework for {\bf Z}ero-{\bf S}hot {\bf L}ink {\bf P}rediction (HNZSLP) that leverages character n-gram information for ZSLP. Our approach works by first constructing a hierarchical n-gram graph from the surface name of relations. Subsequently, a new Transformer-based network models the hierarchical n-gram graph to learn a relation embedding for ZSLP. Experimental results show that our proposed HNZSLP method achieves state-of-the-art performance on two standard ZSLP datasets.\footnote{The code is available here: \url{https://github.com/ToneLi/HNZSLP}
% \href{https://drive.google.com/file/d/1Mro57n-F9P552qW5jPVDQZkdbzOTp26L/view?usp=sharing}{HNZSLP link}.
}

\end{abstract}

%% file: 2_introduction.tex
\section{Introduction}

%The success of these models however depend on their ability to learn expressive representations for entities and relations that preserve the structure of the knowledge graph~\cite{sun2019rotate,chen2020hitter}.
%Learning link prediction models that can deal with unseen relations has thus become of great importance in KG research communities.

% Link prediction models aim to predict relations between entities in knowledge graphs.   Traditional approaches~ \cite{bordes2013translating} assume that all relation types are known in the KG. This assumption is however unrealistic since KGs are inherently incomplete.  To facilitate research, the zero-shot link prediction (ZSLP) task has been introduced to predict unseen relations by leveraging auxiliary information that bridges the gap between seen and unseen relations~\cite{qin2020generative}.  

Link prediction models aim to predict relations between entities in knowledge graphs (KGs). Majority of these methods learn low-dimensional representations of entities and relations (i.e., knowledge graph embeddings (KGE)), which are then used to infer links between entities. Traditional approaches~\cite{bordes2013translating} assume that all relation types are known in the KG. This assumption is however unrealistic since KGs are inherently incomplete. To tackle this issue, the zero-shot link prediction (ZSLP) task has been introduced for identifying unseen relations by leveraging auxiliary information that bridges the gap between seen and unseen relations~\cite{qin2020generative}.  

%The vast majority of LP models nowadays
% leverage the original KG elements to learn low-dimensional representations dubbed Knowledge
% Graph Embeddings, that are subsequently used to infer new facts

% \begin{table}[ht!]
%     \centering
%     \small
%     \begin{tabular}{l|l}
%     & {\bf Relation:} ``{\tt teammate}'' \\ \hline
%       \bf Wikidata  &  A person in one's team\\
%       \bf NELL  & Two athletes are teammates if they play on \\
%       & the same team
%     \end{tabular}
%     \caption{Two sources with different nuances in the definition (or textual description) for the relation {\tt teammate}}.
%     \label{tab:teammate}
% \end{table}

Little previous work exists on ZSLP as the task is relatively new~\cite{qin2020generative,geng2021ontozsl,wang2021structure}. Most efforts focus on using textual features~\cite{qin2020generative,wang2021structure} or ontologies~\cite{geng2021ontozsl} as auxiliary information for the task. Particularly, \citet{wang2021structure} use surface names of entities and relations while \citet{qin2020generative} use the textual descriptions of relations. %Though driven by a common motivation, that is, seen and unseen relations may share textual features for zero-shot learning, using textual descriptions at present may give rise to key issues as we will see shortly.
%for certain relations
%For example, consider the relation ``{\tt teammate}'' and its definition provided by different knowledge bases as shown in Table~\ref{tab:teammate}. The NELL definition may require that the entities in a ``{\tt teammate}'' relationship are athletes if we are to learn a semantic representation on its textual description. On the other hand, Wikidata's definition is more general and covers the definition of ``{\tt teammate}'' more appropriately.
However, these approaches have two main limitations. First, common knowledge graphs such as WordNet~\cite{miller1995wordnet} and FreeBase~\cite{bollacker2008freebase} often do not include textual descriptions of the relations. As such, these need to be obtained from other external sources (e.g., Wikipedia\footnote{https://www.wikipedia.org/}) and are likely to be noisy, leading to poor performance.  Second, manually obtaining such relation descriptions is a labor-intensive and time-consuming process due to the large size of KGs.

Alternatively, \citet{wang2021structure} proposed learning  word representations from the surface name of relations using a pre-trained language model such as RoBERTa~\citep{liu2019roberta}. As surface names are readily available in the KG, this approach is more robust. However, it faces two fundamental weaknesses. First, context is an essential requirement for any text representation method. Surface names on the other hand are represented by short texts, e.g., a relation ``{\tt teammate}'' will have a single word representation observed in training and will therefore have little to no association with an unobserved relation for zero-shot. Second, neural text encoders lack the ability to capture representations for out-of-vocabulary words. This same problem also applies to ``word''-delimited models \cite{qin2020generative} that aim to learn from textual descriptions of relations. In such cases,
the relation representation ability of current methods may be limited significantly, which inadvertently hurts the zero-shot link prediction performance. 

In this paper, we follow a different direction. Instead of simply learning representations from entire words of a relation's surface name, we hypothesize that leveraging character n-grams\footnote{A character n-gram is defined as a contiguous sequence of n characters.} (or n-grams for brevity) information from the relation name will help in generating better representations of unseen relations in zero-shot settings. Models built on subword units (e.g., character n-grams) have the intrinsic ability of generating representations for rare or out-of-vocabulary words~\cite{santos2021morphological}. Inspired by this, we propose a novel Hierarchical N-gram framework for Zero-Shot Link Prediction (HNZSLP) that learns auxiliary information from character n-grams of the surface name of a relation.  %Hence, we hypothesize that leveraging character n-grams will also enable our model to generate better representations of unseen relations in zero-shot settings. 
HNZSLP consists of three main components: (1) a new hierarchical n-gram graph (or n-gram graph for brevity) for representing the relationships between all the character n-grams of a relation; (2) GramTransformer, a new transformer-based~\citep{vaswani2017attention} model for encoding the relation n-gram graph; and (3) a KG Embedding model which adapts prevalent KGE models (e.g., TransE~\cite{bordes2013translating}, DistMult~\cite{yang2014embedding}, TuckER~\cite{balavzevic2019tucker}) to compute a link prediction score between entities in the zero-shot setting. We perform extensive experiments on two standard datasets for zero-shot link prediction demonstrating the superiority of our method over prior state-of-the-art methods.

Our contributions are the following:
\begin{itemize}
	\item We propose HNZSLP, a new framework that uses the character n-gram information from the relation surface name for ZSLP;
	\item We show that our approach outperforms previous state-of-the-art when evaluated on character and byte-level encoders;
	\item We conduct a thorough analysis of our method, including an ablation study, demonstrating the robustness of HNZSLP. 
\end{itemize}

%% file: 3_related_work.tex
\section{Related Work}

%Knowledge representation learning has been widely used in modeling knowledge graph information. TransE \cite{bordes2013translating} embeds relations and entities from symbolic space to complex vector space.
%%Chen---embeds,...from--to--
%TuckER \cite{balavzevic2019tucker} utilizes the tucker decomposition \cite{malik2018low} to build the connection between different knowledge graph triples. In the testing process, these models infer the missing links by exiting relations or entities.

%Subsequently, there are many existing works that focus on inferring the missing elements by using the language model, such as \cite{yao2019kg,liu2020k}.
%These methods can predict relations by simple vector operations. However, unfortunately, 

%Knowledge representation learning has been widely used in modeling knowledge graph information. raditional

\subsection{Link Prediction}
So far, a variety of works have been proposed for link prediction, and the difference in their architecture ranges from the scoring function to how the optimization problem is modeled to learn entities and relation embeddings. As current work is vast and fast growing, we restrict ourselves to reviewing only those closely related to our work. Some of the well-known methods include the translation-based model TransE~\cite{bordes2013translating}, which requires that the tail entity embedding is close to the sum of the head and relation embeddings; the bilinear model DistMult~\cite{yang2014embedding} that interprets the relation as a bilinear map and uses multiplicative interactions to learn entity and relation embeddings; the non-bilinear model TuckER \cite{balavzevic2019tucker} utilizes the tucker decomposition \cite{malik2018low} to build the connection between different knowledge graph triples. Although performance has been achieved incrementally, these approaches in their original form are unable to learn embeddings for unseen relations. This is due to the fact that they learn entities and relation embeddings using the topological structure of the KG. We refer the reader to the work by \cite{rossi2021knowledge} for further background on such methods.

%Traditional approaches~\cite{bordes2013translating,balavzevic2019tucker} in link prediction are only able to predict relations that are observed during training but are unable to deal with newly-add relations in KGs. This needs to be expanded a little bit...Any work from 2020-21-22? Any work from 2013 to 2019? Add references

\subsection{Zero-shot Link Prediction}
The zero-shot link prediction~\cite{qin2020generative} is a new task that aims to predict unseen relations between entities by using auxiliary information to bridge the gap between seen and unseen relations. \citet{qin2020generative} uses textual information of the relation as auxiliary information and applies a Zero-Shot Generative Adversarial Network (ZSGAN) to learn the unseen relation embedding for the task. An Ontology-enhanced Zero-Shot Learning (OntoZSL)~\cite{geng2021ontozsl} obtains structural information of relations from the ontology and combines it with the textual descriptions of the relations for zero-shot learning. 
Despite the success, these textual descriptions are typically not present in knowledge graphs and therefore these methods rely on external sources to collect such data. This makes it labor-intensive and time-consuming to obtain the most representative descriptions of entities and relations.

\subsection{Character-level Information for Zero-shot Learning}
An emerging trend is to use the character-level information of the raw text in zero-shot learning. Byt5~\cite{xue2021byt5} is one of such models that uses a language model T5~\cite{raffel2019exploring} to process byte or character sequences. Charformer~\cite{tay2021charformer} improves upon Byt5 by introducing a gradient-based subword tokenization module to learn the character information. Our proposed model is somewhat aligned with these models in the sense that we consider the character information in the text. However, we consider the process of how words are formed. That is, being considered as a sequential combination of characters/n-grams or a hierarchical structure, whereby different n-grams aggregate up to a complete word. We model this structure, referred to as a n-gram graph structure, using a self-attention based Transformer~\cite{vaswani2017attention} due to its success in graph learning~\cite{ahmad2021gate,cai2020graph,lyu2021let,yao2020heterogeneous}.

%% file: 4_preliminaries.tex
\section{Problem Statement}
\label{con:task defination}

% A Knowledge Graph (KG) is a directed graph that consists of entities and relations. Formally, we define 

A Knowledge Graph (KG) is defined as a graph $\mathcal{G}=(\mathcal{R},\mathcal{E})$, where $\mathcal{E}$ denotes a set of entities and $\mathcal{R}$ denotes the set of relations among these entities. In a KG, the entities and relations are usually organized as {\em facts} and each fact is defined as a triplet $(h, r, t)$ where $h, t\in \mathcal{E}$ and $r \in \mathcal{R}$ denote the head entity, tail entity and the relation between the two entities, respectively. 

% \textbf{ZSL Link Prediction:} 
% In zero-shot link prediction, there are two relation sets, a seen relation set $R_s$ and an unseen relation set $R_u$ respectively, where $R_s \cap R_u =\emptyset$. The training set $T_{s}=\{(h_{},r_{s},t, C_{h_{},r_{s}})|h \in E, r_{s} \in R_{s}, t \in E\}$ is built with each seen relation $r_{s}$.  $C_{h,r_{s}}$ denotes a candidate set corresponding to the query tuple $(h_{},r_{s})$.
% Meanwhile, a testing set can be formulated as $T_{u}=\{(h_{},r_{u},t,C_{h_{},r_{u}})|h \in E, r_{u} \in R_{u}, t \in E\}$, where  $t$ is the ground-truth tail entity. Given a newly added relation $r_{u} \in R_u$, the objective of our model is to assign the highest ranking score to the ground-truth tail entity $t$ which is included in the candidate set $C_{h,r_{u}}$.
In the zero-shot link prediction problem, we assume that there are two disjoint relation sets in the KG, a seen relation set $\mathcal{R}_s$ and an unseen relation set $\mathcal{R}_u$, where $\mathcal{R}_s \cap \mathcal{R}_u =\emptyset$. We are given a training set $\mathcal{D}_{s}=\{(h,r_{s},t)|h,t \in \mathcal{E}, r_{s} \in \mathcal{R}_{s}\}$ in which the facts are involved with observed relations $r_{s} \in \mathcal{R}_{s}$. Meanwhile, we define the test set as $\mathcal{D}_{u}=\{(h,r_{u},t,C_{(h,r_{u})})|h, t \in \mathcal{E}, C_{(h,r_{u})} \subseteq \mathcal{E}, r_{u} \in \mathcal{R}_{u}\}$, where $t$ is the ground-truth tail entity and $C_{(h,r_{u})}$ denotes a candidate set corresponding to a query $(h,r_{u})$. Given a query $(h, r_u)$, the objective of zero-shot link prediction is to find the ground-truth tail entity $t$ from the candidate set $C_{(h,r_{u})}$.

%% file: 5_method.tex
\section{HNZSLP}
\begin{figure*}[htbp]
	\centering
	\includegraphics[width=1.5\columnwidth]{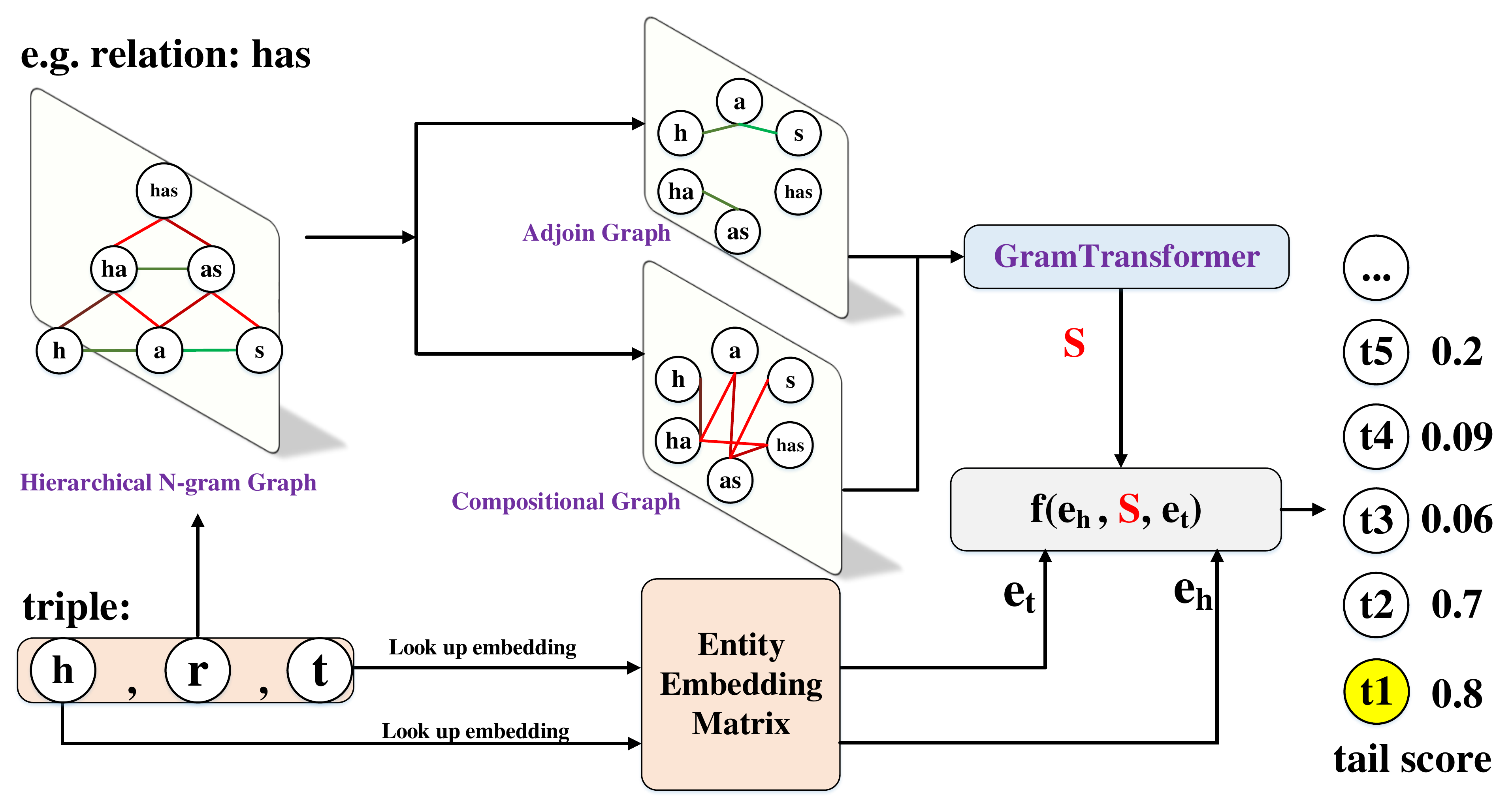} % Reduce the figure size so that it is slightly narrower than the column. Don't use precise values for figure width.This setup will avoid overfull boxes.
	\caption{Overview of HNZSLP, we take a relation 
	\textit{has} as an example. In the adjoin Graph, the green line denotes the adjoin relationship. In the compositional Graph, the red line indicates the compositional relationship. In these two graphs, different color depths represent different attention weights. The node in the adjoin graph and the compositional graph is called  neighbor node and superior node separately.}.
	\label{con:HNZSLP}
\end{figure*}

Figure \ref{con:HNZSLP}  gives an overview of our HNZSLP framework, which consists of three major parts: (1)  Hierarchical N-gram Graph Constructor constructs a hierarchical n-gram graph from each relation surface name, where the n-gram graph can be further decomposed into an {\it adjoin} graph and {\it compositional} graph to simplify the learning of the relation representation; (2) GramTransformer constructs the relation representation by modeling over the adjoin and compositional graphs; (3) KG Embedding Learning Module combines the embeddings of the head entity and relation to predict the tail entity and update the embeddings.

\subsection{Hierarchical N-gram Graph Constructor} \label{con:detail of lattice graph}

% Our lattice graph consists of nodes and relations which denote the connection for each two nodes. Next, we introduce the nodes and relations in details.

% \subsubsection{Nodes:} 
\paragraph{Node Construction} For each word token in the relation surface name, we first collect all possible $n$-grams, where $n$ is valued from $1$ up to the maximum gram size $M$ of a word. For example, $M=3$ for the relation \textit{has} in Figure~\ref{con:HNZSLP}. All n-grams are treated as nodes in the hierarchical n-gram graph. Suppose the relation surface name contains multiple words, the n-grams of each word are composed into a unified n-gram graph.
%%%%%%%%%
For each hierarchical n-gram graph, we denote all its nodes as a sequence $X=\{x_1, x_2, \cdots, x_b\}$, where $b=\frac{M(M+1)}{2}$. Let $\mathbf{X}=\{\mathbf{x}_1, \mathbf{x}_2, \cdots, \mathbf{x}_b\}$ be the corresponding node embeddings for the graph.

% For each hierarchical n-gram graph, we denote all its nodes as a sequence $X=\{x_1, x_2, \cdots, x_b\}$, where $b=\frac{M(M+1)}{2}$. For each node $x_i$, we define $\mathbf{x}_i$ as its embedding. Then, we define a matrix $\mathbf{X}=\{\mathbf{x}_1, \mathbf{x}_2, \cdots, \mathbf{x}_b\}$ to represent all concatenated node embeddings for the hierarchical n-gram graph.

% In order to learn the embedding of each gram, we use a random vector matrix $ \mathbb{R}^{b \times d}$ to initialize grams  that project each gram to a single gram representation $X_{i} \in \mathbb{R}^{b \times d} $.  The n-gram graph nodes $N(G)$ and the node embedding  matrix $V(G)$ can be defined as:
% $$N(G)={x_1, x_2,\cdots,x_b}, V^n(G)={X_1, X_2,\cdots,X_b}$$
% where we construct each gram $x_i$ and its embedding $X_i$ for $i \in 1,...,b$, and $b$ is the size of n-grams for each word. In  Figure~\ref{con:HNZSLP}, for relation word "has", $b=6$. %For easy to train, when the word length is bigger than $n$ , we replace the $n$-th gram blocks with word which as the final gram level, such as in Figure~\ref{con:HNZSLP}, the $3$-th gram blocks of word "part" are "par" and "art", we replace the ("par" and "art")  with "part". This way can help receive word information.

%\textbf{note:} each word form its self lattice graph, for many words in a relation, we set consider the gram1, gram2,... 

% \subsubsection{Relations:}
\paragraph{Edge Construction}
We define two types of edges among n-grams in the hierarchical n-gram graph: \emph{adjoin} edge and \emph{compositional} edge. The adjoin edge implies that two n-grams at the same hierarchical level are neighbors, e.g., the edge between nodes ``h'' and ``a'' in Figure~\ref{con:HNZSLP}. The compositional edge implies that the n-gram node at a higher-level (i.e., a superior node) is a composition of the adjacent n-gram nodes at the immediate lower-level, e.g., the edge between node ``h'' and ``ha'', and ``a'' and ``ha'' in Figure~\ref{con:HNZSLP}. According to the two edge types, we can decompose the n-gram graph into the adjoin graph and compositional graph, as shown in Figure \ref{con:HNZSLP}.

% The compositional edge implies that the lower-level n-gram is a portion of the higher-level n-gram, e.g., the relation between node ``h'' and ``ha'' in Figure~\ref{con:HNZSLP}. According to the two relation type, we can decompose the n-gram graph into the adjoin graph and compositional graph, as the example shown in Figure \ref{con:HNZSLP}.
% To obtain the embeddings of the adjoin and compositional edges, we first collect their textual definitions from

For the adjoin and compositional edges, we first define  their textual definitions based on Wikidata,\footnote{https://www.wikidata.org/wiki/Wikidata:Main\_Page} and calculate their embeddings using Sentence-BERT~\cite{reimers2019sentence}. For later use, we define the embeddings of the adjoin and compositional edges as $\mathbf{r}_a \in \mathbb{R}^{d}$ and $\mathbf{r}_c \in \mathbb{R}^{d}$, respectively.  

Note that some surface names of the relations may be a long sequence of words, which may result in a large set of nodes in the n-gram graph, making it hard to process.
% in the when the length of word is big or the number of word in a relation is big (such as relation "concept:agricultural product growing in state or province" in NELL), the n-gram graph will become very big, it is hard for machine to progress this graph.  
To boost the graph construction process, we reduce the number of nodes in the hierarchical n-gram graph using two strategies (see details in Appendix Section \ref{Hierarchical N-gram Graph building and node selection}). 
% For example, the listed nodes of relation "a part of" are (a,p,a,r,t,o,f | pa,ar,rt,of | par,art| part).

\subsection{GramTransformer} 

% \subsubsection{Overview}

\begin{figure}[t]
	\centering
	\includegraphics[width=1\columnwidth]{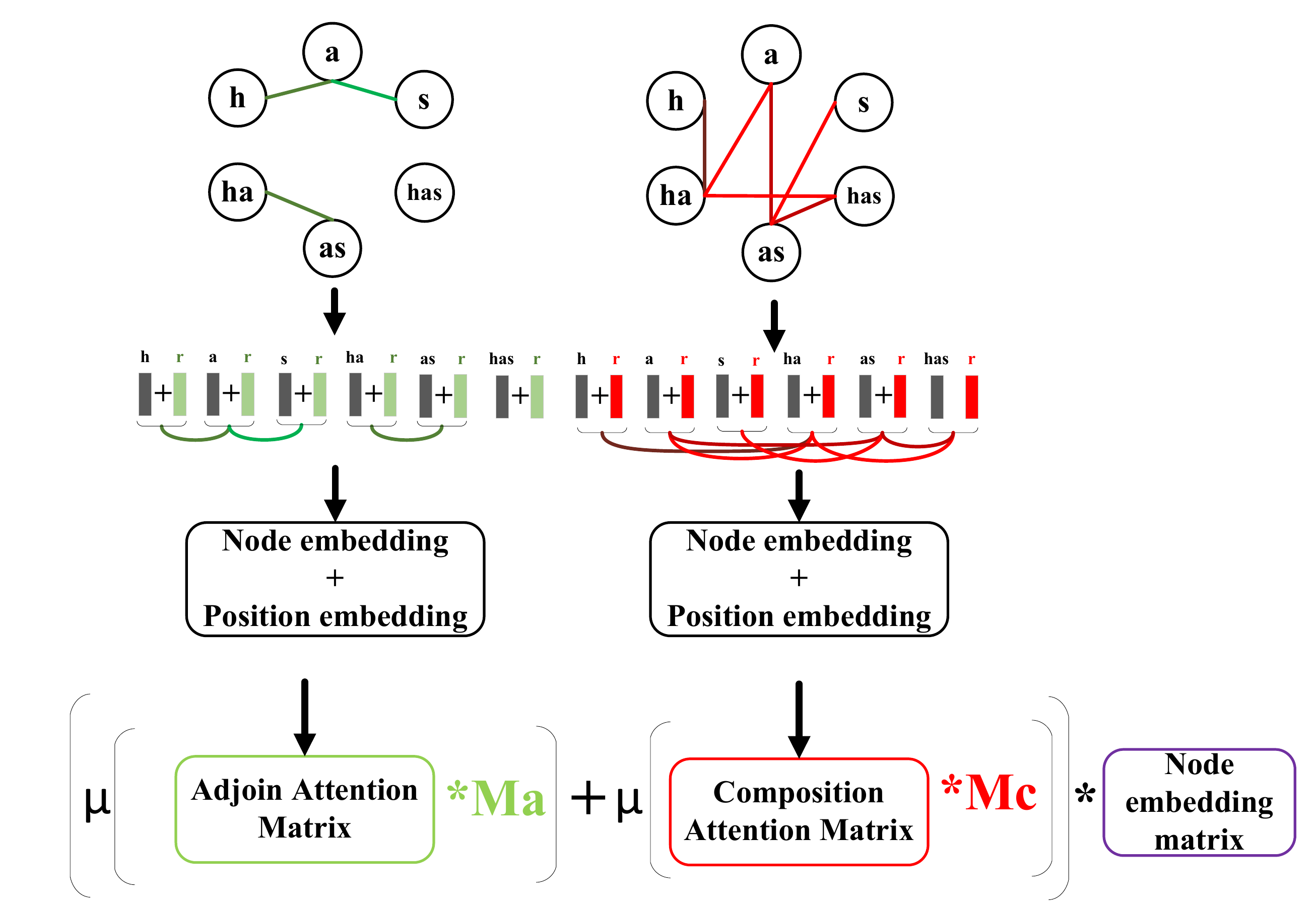} 
	\caption{Overview of the relation enhanced mask attention used in the GramTransformer. $M_a$ and $M_c$  are the mask matrix that models the relationship among nodes in the adjoin and compositional graphs, respectively. $\mu$ is the softmax function.
	The green rectangle refers to the embedding $\mathbf{r}_a \in \mathbb{R}^{d}$. 	The red rectangle refers to the embedding $\mathbf{r}_c \in \mathbb{R}^{d}$.}
	\label{con:HNZSLP_attention}
\end{figure}

%denotes the connection of adjoin graph

% Given a graph, current graph neural networks calculate the node representation based on all the first-hop relations and neighbor nodes. All nodes or relations will be mapped into an same embedding space. 
% Considering that our hierarchical n-gram consists of an adjoin graph and a compositional graph. It is not suit for our n-gram graph with adjoin nodes, composition nodes and two types relations. So we introduce a new model, named as GramTransformer, which provides an aggressively different paradigm that compute the relation and node information.
We propose a GramTransformer to efficiently extract hierarchical n-gram features. 
%%%%%%%%%
The difference between the standard Transformer and our GramTransformer is the attention calculation, as shown in Figure~\ref{con:HNZSLP_attention}. The most important characteristic of the GramTransformer is that it can encode the neighbor node and superior node information while taking the edge information in the adjoin and compositional graphs into account.

In this way, a node can directly learn the information from different neighborhoods. These operations are achieved by our proposed relation enhanced mask attention mechanism. Specifically, we decompose the original n-gram graph into the adjoin graph and compositional graph based on adjoin edge and compositional edge. We then initialize the node embeddings in each subgraph as the sum of the node embedding, edge embedding $\mathbf{r}_a$ (or $\mathbf{r}_c$), and position \footnote{for the position about nodes, please refer Appendix Section \ref{Hierarchical N-gram Graph building and node selection}} embedding  \cite{vaswani2017attention}. Next, the initialized subgraph elements with mask matrix is used to learn the attention matrix, highlighting the relevant features of the n-gram graph. Multiple layers of relation enhanced mask attention networks are stacked to calculate the final node representation. At each layer, a node vector is updated based on neighbor nodes and associated edge types. The nodes' vectors at the last layer is considered as the final n-gram graph representation.

%Next, the subgraph with mask matrix is used to learn the attention matrix for the relevant graph.

\subsubsection{N-gram Graph Encoder}
Our graph encoder aims to transform an input n-gram graph into a set of node embeddings. To calculate the node information in our graph, the central problem is how to calculate the node vectors based on the different subgraphs (adjoin graph or compositional graph). To this end, we propose a relation enhanced mask attention mechanism, which is an extension of the self-attention mechanism to relate the different nodes across the subgraphs.

To maintain the graph structure information, our idea is to introduce the explicit edge information and incorporate it into the subgraph attention score computation.
% As suggested in \cite{cai2020graph}, our treatment of explicit edge information in different subgraphs also brings other advantages, such as some work \cite{guo2019densely,zhang2020relational} regard the edges as the nodes, and end up sharing the same semantic space, which is not ideal \cite{cai2020graph}.
% The idea is to first incorporate the edge and node information in each subgraph  and learn the attention information separately. 
We introduce the mask matrix to model the relationship among nodes in each subgraph.   Recall,  for the standard self-attention, an L-layer Transformer takes $\mathbf{X}$ as input and produces the latent representation $\mathbf{H}^l=(\mathbf{h}_1^l,\mathbf{h}_2^l...,\mathbf{h}_b^l)$ of relations. To enhance the semantic representation of the input, multi-head self-attention is used in each Transformer Layer. Specifically, the output of $(l-1)^{\rm th}$ Transformer layer is projected to a query matrix $\mathbf{Q}_{l}$ and a set of key-value $(\mathbf{K}_{l}, \mathbf{V}_{l})$ pairs, $$\mathbf{Q}_l=\mathbf{H}^{l-1} \mathbf{W}_l^{q}, \mathbf{K}_l=\mathbf{H}^{l-1}\mathbf{W}_l^{k}, \mathbf{V}_l=\mathbf{H}^{l-1}\mathbf{W}_l^{v}$$

\noindent where $\mathbf{W}_l^q, \mathbf{W}_l^k, \mathbf{W}_l^v\in \mathbb{R}^{d_{model} \times d_k}$ denote the learnable weight matrices. $d_{model}$ is the model dimension, $d_k$ is the head dimension. The output of a self-attention head $\overline{\mathbf{H}}^{l}$ is calculated by:
%Adaptively weighted attention matrix is calculated as follow.
\begin{align}
% \label{Attention}
	&\overline{\mathbf{H}}^{l}=\mu(\frac{\mathbf{Q}_{l}\mathbf{K}_{l}^\textup{T}}{\sqrt{d_k}} )\mathbf{V}_{l}
	\label{A_h}
\end{align}

where $\mu$ denotes the $softmax$ function. The self-attention learns the implicit relationships between nodes in the hierarchical n-gram graph. 

\subsubsection{Relation Enhanced Mask Attention}
As our n-gram graph consists of adjoin graph and compositional graph, we respectively compute the attention head as follows,

\begin{align}
\label{Attention}
    &\mathbf{Q}_{l}^a=\mathbf{K}_{l}^a=\mathbf{V}_{l}^a= \mathbf{Q}_{l}+\mathbf{r}_a \\
    &\mathbf{Q}_{l}^c=\mathbf{K}_{l}^c= \mathbf{V}_{l}^c=\mathbf{Q}_{l}+\mathbf{r}_c\\
	&\tilde{\mathbf{H}}^{l}=[\mu(\frac{\mathbf{Q}_{l}^a(\mathbf{K}_{l}^a)^\textup{T}}{\sqrt{d_k}})+\mu(\frac{\mathbf{Q}_{l}^c(\mathbf{K}_{l}^c)^\textup{T}}{\sqrt{d_k}})]\mathbf{V}_{l}
	\label{con:relation_types}
\end{align}

%Each attention term in Eq.\ref{con:relation_types} corresponds to different meanings.

where we split the node embedding $Q_l$ into neighbor node embedding $Q^a_l$ and superior node embedding $Q^c_l$. Then we compute the attention score based on the embeddings of the nodes and edges in the subgraph. In Eq.\ref{con:relation_types}, the first term represents the node weight calculated from its adjoin neighbors. The second term represents the node weight calculated using the compositional edge information. In comparison, our model can calculate the node embedding respectively based on the different edge types, it can compute the node embedding more precisely than the standard mask self-attention which does not consider the edge type. The comparative experiment is in our Section \ref{Ablation Experiments}.

% which maps the all types relation in the same vector space.

To denote the node connection in subgraphs, the central idea is to incorporate the mask matrix into the attention matrix of self-attention, which can impose the structure of the n-gram graph and reassign the attention weight for each relation.

We denote the mask matrix $\mathbf{M} \in \mathbb{R}^{m \times m}$, where $\mathbf{M}_{ij}\in[0,1]$ denotes the connection between node at position $i$ and $j$ in the input n-gram node list. $1$ denotes there is a connection between two nodes. So our proposed attention strategy can be redefined as,
\begin{align}
% \label{Attention}
	&\mathbf{H}^{l}=[\mu(\frac{\mathbf{Q}_{l}^a(\mathbf{K}_{l}^a)^\textup{T}}{\sqrt{d_k}})  {\mathbf{M}_a}+\mu(\frac{\mathbf{Q}_{l}^c(\mathbf{K}_{l}^c)^\textup{T}}{\sqrt{d_k}})\mathbf{M}_c]\mathbf{V}_{l}
	\label{ada_wei}
\end{align}
where $\mathbf{M}_a$ indicates the relationship among nodes in the adjoin graph, $\mathbf{M}_c$ indicates the relationship among nodes in  the compositional graph.
%%%%%%%
Hence, for an input n-gram graph, our graph encoder module produces the attention head $\mathbf{H}^{l}$ which is fed to subsequent layers of the GramTransformer to output the node representations $\mathbf{h}_1^l,\mathbf{h}_2^l...,\mathbf{h}_b^l$. We then apply a mean pooling over these representations to obtain the relation embedding $\mathbf{S}$ of the surface name.

\subsection{Embedding Learning Module} 

We randomly initialize entity embedding matrix $\mathbf{E} \in \mathbb{R}^{|\mathcal{E}| \times d^e} $, where each row vector is the embedding of an entity and $d^e$ is the dimension of the entity embedding. In a triplet $(h, r, t)$, we define $\mathbf{e}_h$ and $\mathbf{e}_t$ as the embedding of the head entity and tail entity retrieved from the embedding matrix $\mathbf{E}$.
Given the entity embeddings $\mathbf{e}_h$ and $\mathbf{e}_t$, and the relation embedding $\mathbf{S}$ computed by our proposed GramTransformer, we then define a scoring function $f(\cdot)$ that assigns a score $\eta$ to each triple $(h,r,t)$,
\begin{align}
	&\eta=f(
	\mathbf{e}_h,\mathbf{S},\mathbf{e}_t)  \label{con:score}
\end{align}
where the scoring function $f$ can be replaced by any knowledge graph embedding model, e.g., TransE, DistMult. The model is trained with cross-entropy loss.

% where the scoring function $f$ can be replaced by any knowledge graph embedding models, e.g., TransE\footnote{The score function of TransE is defined as: f(e_h,S,e_t) = -	\left \|e_h+S-e_t\right \|} and DistMult\footnote{The score function of DistMult is defined as: f(e_h,S,e_t) = 	$\left \langle e_h,S,e_t\right \rangle$, where 	$\left \langle .\right \rangle$ denotes the generalized dot product.}. The model is trained with cross-entropy loss. 

% The embedding module in the existing works~\cite{qin2020generative,geng2021ontozsl} are pre-trained on ZSL knowledge embedding and the embedding of the entities are fixed during training. In this paper, as the lattice information is taken into account and it is expected that the embedding of entities should be updated based on the structure of the lattice graph. To that end, our embedding module allows the embedding of the entities to be updated during training so that these learned entity embeddings together with the lattice graph can further improve the prediction performance on the unseen relations.

% \subsection{Tail Entity Inference} 

During the inference process, the trained HNZSLP scores each candidate tail entity $t^{'} \in C_{(h,r_u)}$ given the
query $(h, r_u)$. Let $\mathbf{S}_u$ be the embedding of $r_u$ computed by GramTrasformer, the entity with the highest score in the candidate entity set is selected as the predicted tail entity:
\begin{align}
	t^{*}=\mathop{\arg\max}\limits_{t^{'} \in C_{(h,r_u)}}f(\mathbf{e}_h,\mathbf{S}_u, \mathbf{e}_{t^{'}})
\end{align}
where $t^{*}$ refers to the predicted tail entity.

%% file: 6_experiment.tex
\section{Experiments}
We validate HNZSLP by comparing HNZSLP's performance with several recent works, including ZSGAN~\cite{qin2020generative} and OntoZSL~\cite{geng2021ontozsl}. ZSGAN exploits the generated description embeddings of unseen relations to predict the tail entity while OntoZSL introduces the ontology strategy in the task. Following ~\citet{geng2021ontozsl}, we further compare with the baselines ZSL-TransE and ZSL-DistMult that use Word2vec \cite{vinyals2015neural}, and respectively  employ TransE~\cite{bordes2013translating} and DisMult~\cite{yang2014embedding} as KGE models for zero-shot link prediction. For a fair comparison, we exclude the results of \cite{wang2021structure} as this approach does not show the results in our used datasets. We use four commonly used metrics, mean reciprocal ranking (MRR), hits@\textit{10}, hits@\textit{5}, hits@\textit{1}, and evaluate on two benchmark datasets, including NELL-ZS and Wikidata-ZS (Wiki-ZS) proposed by \citet{qin2020generative}. A summary of dataset statistics is given in Table \ref{con:ZSL data}.

\begin{table}[ht]
	\centering
	
	\renewcommand\arraystretch{1.3}
	\scalebox{0.7}{
	\begin{tabular} {cccc}
		\hline 
		
		\hline	
		Dataset& \# Entities & \# Triples &   \# Train/ \# Dev/ \# Test\\ 
		\hline		
		NELL-ZS&65,567 &  188,392& 139/\,10/\,32\\
		Wiki-ZS&605,812& 724,967 &469/\,20/\,48\\
		\hline
	\end{tabular}}
		\caption{ Datasets Statistics, column 4 refers to the number of relations in the different set.}
	\label{con:ZSL data}
\end{table}

\subsection{Implementation details}

On NELL-ZS (or Wiki-ZS), each word in the surface name of  relation is set to $13$-gram (or $15$-gram) and the number of nodes in the n-gram graph is set to $90$ (or $70$). Each node in the n-gram graph is randomly initialized with a 200-dim embedding. The GramTransformer contains one multi-head attention block with three attention heads and a 200-dim feed-forward layer. The dropout rate in the multi-head attention and feed-forward layer is set to $0.5$.
Entities are also randomly initialized with 200-dim embeddings. During training, we use Adam~\cite{kingma2014adam} as the optimizer and a Cross-entropy is used as the loss function with a learning rate of $0.0005$. We use label smoothing to prevent the model from becoming over-confident. All embeddings are fine-tuned during training.

\begin{table*}[ht]
	\centering
		
	\renewcommand\arraystretch{1.3}
	\scalebox{0.8}{
	\begin{tabular} {c|c|cccc|cccc}
		\hline 
		\multicolumn{2}{c}  {}&\multicolumn{4}{c}  {NELL-ZS}& \multicolumn{4}{c}  {Wiki-ZS} \\
		\hline
		KGE model&Method& MRR &  hits@\textit{10} &  hits@\textit{5}&hits@\textit{1} &MRR &  hits@\textit{10} &  hits@\textit{5}&hits@\textit{1}  \\ 
		\hline	
	
		\multirow{4}*{TransE}&ZSL-TransE\cite{qin2020generative}&0.097&0.203&0.147&0.043 & 0.053&0.119&0.081&0.018\\
		
		%~&Cosine-Map&0.097&0.212&0.134&  0.042& 0.084&0.198&0.109&0.036\\
		~&OntoZSL\cite{geng2021ontozsl} & 0.250&0.399& 0.327& 0.172& 0.184&0.265&0.215&0.138 \\
		~&ZSGAN\cite{qin2020generative}&0.234 &0.373 & 0.304 & 0.160& 0.177&0.258&0.207&0.131 \\
		~&HNZSLP&\textbf{0.289} &\textbf{0.413}&\textbf{0.359}&\textbf{0.222}&\textbf{0.252} &\textbf{0.307}&\textbf{0.281}&\textbf{0.219}\\

		\hline
		\multirow{4}*{DistMult}&ZSL-DistMult\cite{qin2020generative}&0.235 &0.326&0.284&  0.185& 0.189&0.236&0.210&0.161\\
		
		%~&Cosine-Map&0.088 &0.179&0.111&  0.045& 0.089&0.197 &0.107&0.040\\
		~&OntoZSL\cite{geng2021ontozsl}& 0.256&\textbf{0.385}&0.318 &0.188& 0.211 & \textbf{0.289}& 0.238&0.167\\
		~&ZSGAN\cite{qin2020generative}&0.249 &0.376 & 0.306 & 0.183& 0.207 & 0.284 & 0.235&0.164\\ 
		~&HNZSLP& \textbf{0.276}& 0.383 &\textbf{0.333}&\textbf{0.216}& \textbf{0.232}& 0.279&\textbf{0.254}&\textbf{0.204}\\
		\hline
	\end{tabular}}
	\caption{Zero-shot link prediction results in NELL-ZS and Wiki-ZS. The baseline results  were obtained from \cite{geng2021ontozsl}. The KGE models in the first column correspond to $f(.)$ in equation \ref{con:score}.}
\label{con:link-prediction-result}
\end{table*}

\subsection{Main Results} 
Evaluation results are shown in Table \ref{con:link-prediction-result}. Results are the average over 5 runs.  We find that our approach outperforms all previous methods on the different KGE models. With TransE and DistMult, our model achieves  hits@1 scores of 0.222 and 0.216 respectively on NELL-ZS, outperforming the previous best-performing network OntoZSL by a margin of 0.05 and 0.028.  With DistMult, we also find that our model achieves the best performance on hits@1, but slightly underperforms the best-performing network OntoZSL on NELL-ZS. It is worth noting that OntoZSL utilizes external ontology resources, thus, they present an additional advantage over ours that do not consider external knowledge. In real applications, ontology is not always available, which limits the scalability of their method.
%%%%%%%%
On Wiki-ZS, our model also sets a new hits@1 score. Particularly, for TransE, we improve over the state-of-the-art OntoZSL by about 0.081 points. Similar performance is achieved on other metrics for the different KGE models. These results indicate that sufficient information is contained in the relation surface name to achieve zero-shot link prediction, and our proposed method is effective, it can utilize the n-gram graph to transfer the knowledge between seen relation and unseen relation is effective.

\section{More Analysis}

\begin{table*}[ht]
	\centering
		
	\renewcommand\arraystretch{1.3}
	\scalebox{0.8}{
	\begin{tabular} {c|c|cccc|cccc}
		\hline 
		\multicolumn{2}{c}  {}&\multicolumn{4}{c}  {NELL-ZS}& \multicolumn{4}{c}  {Wiki-ZS} \\
		\hline
		KGE model&Method& MRR &  hits@\textit{10} &  hits@\textit{5}&hits@\textit{1} &MRR &  hits@\textit{10} &  hits@\textit{5}&hits@\textit{1}  \\ 
		\hline	
	
		\multirow{6}*{TransE}&ZSL-ByT5 $\triangledown$  &0.103 &0.173&0.129&  0.064& 0.110&0.193&0.134&0.064\\
		~&ZSL-CharFormer $\triangledown$ &0.269 & 0.383 &0.337 & 0.202 & 0.170 &0.236 &0.201  &  0.132\\
		
	    ~&OntoZSL\cite{geng2021ontozsl} & 0.250&0.399& 0.327& 0.172& 0.184&0.265&0.215&0.138 \\
		~&HNZSLP-WNG $\circ$ & 0.278&0.409&0.350& 0.203&0.231 &0.273&0.262&0.194\\
		~&HNZSLP-WG $\circ$&0.244  &0.382&0.322& 0.173& 0.232&0.284&0.258&0.199\\
		
		~&HNZSLP&\textbf{0.289} &\textbf{0.413}&\textbf{0.359}&\textbf{0.222}&\textbf{0.252} &\textbf{0.307}&\textbf{0.281}&\textbf{0.219}\\
	
		\hline
		\multirow{6}*{DistMult}&ZSL-ByT5 $\triangledown$  &0.233 &0.382&0.317&  0.150& 0.196&0.255&0.232&0.160\\
		~&ZSL-CharFormer $\triangledown$ &0.251&0.380&0.320&0.183& 0.205&0.270&0.251&0.153\\
	~&OntoZSL\cite{geng2021ontozsl}& 0.256&\textbf{0.385}&0.318 &0.188& 0.211 & \textbf{0.289}& 0.238&0.167\\
			~&HNZSLP-WNG $\circ$& 0.274 &0.381&0.330&0.195 & 0.178&0.208&0.192&0.157\\
		~&HNZSLP-WG $\circ$ & 0.259 &0.372&0.321& 0.195& 0.221&0.264&0.242&0.197\\
		
		~&HNZSLP& \textbf{0.276}& 0.383 &\textbf{0.333}&\textbf{0.216}&\textbf{ 0.232}& 0.279&\textbf{0.254}&\textbf{0.204}\\

		\hline
	\end{tabular}
	
	}
	\caption{The performance of HNZSLP with different character/byte learning models (denote as $\triangledown$) and its variants (denote as $\circ$). WNG refers to without n-gram graph, WG refers to without GramTransformer. }
\label{con:alaby_study}
\end{table*}

%Through the above analysis, we can find the overall results show a better benefit of HNZSLP over existing methods. 

In order to further explore the effectiveness of our framework, we perform a series of analyses based on different characteristics of our model. First, we explore the effectiveness of our proposed GramTransformer with two latest works that learn text information from the character level or byte level. The contribution of our model components can also be learned from ablated models. So we propose two model variants to help us validate the advantages of the n-gram graph information and GramTransformer. 
% utilize GramTransformer to calculate the n-gram graph information further. 
Next, we explore the performance of our model with a different number of nodes in the n-gram graph. In the last,
we did a comparison with the method which applies the language model in this task.

For more experiments about the HNZSLP performance 
in the OOV problem, the impact of different node selection strategies, please refer to  Appendix Sections \ref{out-of-vocalulary}, \ref{Node Selection Strategies}.

\subsection{Comparison with Character/Byte Models}
% To quantitatively evaluate the effectiveness of GramTransformer, we compare the performance of  HNZSLP against we proposed two models (\textbf{ZSL-CharFormer} and \textbf{ZSL-ByT5}) which are based on the latest work about character learning and byte learning. These two models utilize  CharFormer \cite{tay2021charformer} and ByT5 \cite{xue2021byt5} to  calculate the embedding $S$ of relation surface name in  eq.\ref{con:score} separately. They are fine-tuned in the dev dataset. 

% To evaluate the effectiveness of our proposed GramTransformer, we compare 

% the performance of  HNZSLP against we proposed two models (\textbf{ZSL-CharFormer} and \textbf{ZSL-ByT5}) which are based on the latest work about character learning and byte learning. These two models utilize  CharFormer \cite{tay2021charformer} and ByT5 \cite{xue2021byt5} to  calculate the embedding $S$ of relation surface name in  eq.\ref{con:score} separately. They are fine-tuned in the dev dataset. 

To evaluate the effectiveness of our proposed GramTransformer, we explore two state-of-the-art methods for character/byte level learning, including CharFormer \cite{tay2021charformer} and ByT5 \cite{xue2021byt5} to calculate the relation surface name embedding $S$ (as shown in \eqref{con:score}) at the character or byte level, respectively. Accordingly, we propose \textbf{ZSL-CharFormer} and \textbf{ZSL-ByT5} for ZSLP using the same experimental setup as our method for a fair comparison with our method.
%%%%%%%%%%
Table~\ref{con:alaby_study} shows the performance comparison. On the dataset NELL-ZS, our model can achieve the hits@1 score of 0.222, outperforming the best model ZSL-CharFormer by a large margin of 0.02 hits@1 score using the TransE KGE model. On the dataset Wiki-ZS, our model also outperforms the best model ZSL-CharFormer by an impressive margin of 0.087 on the same TransE KGE model. The advantages of our model are also verified by MRR, hits@10, and hits@5. Otherwise, the model performance under different KGE models also is compared.
%%%%%%%%%%

In CharFormer, \citet{tay2021charformer} lists a fixed number of subword blocks and uses an attention-based method to choose the best subword at each character position. By utilizing the 
stride window to get the subwords, this work ignores the semantic information of other subwords outside the window. Meanwhile, in our work, we propose to use n-gram to help calculate the relation information. Our approach is much more conducive to preserve semantic information as it considers the effective modeling of rare words or OOV words. Again, our results demonstrate that the hierarchical n-gram graph information is important to express the semantic information of the relation.

\subsection{Ablation Experiments}
\label{Ablation Experiments}
% \subsection{Influence of n-gram Information and Relation type}
The contribution of our model components can also be learned from ablated models. We introduce two ablated models of HNZSLP, (1) {\bf HNZSLP-WNG} uses a traditional Transformer to learn the information of relation surface name from the 1-gram level; (2) {\bf HNZSLP-WG} uses the standard self-attention to learn the information of n-gram graph, instead of our proposed  GramTransformer that incorporates different sub-graph information. We find that the performance of HNZSLP degrades as we remove important model components. Specifically, both HNZSLP-WNG and HNZSLP-WG perform poorly when compared to HNZSLP, indicating the importance of modeling the information of the n-gram graph.

% We find that the performance of HNZSLP degrades as we remove important model components. Specifically, {\bf HNZSLP-WNG} just uses a traditional Transformer to learn the information of relation surface name from the 1-gram level. This method performs poorly as compared to HNZSLP, indicating the importance of modeling the information of the n-gram graph. We also find that HNZSLP-WG uses the standard self-attention to learn the information of n-gram graph, instead of our proposed  GramTransformer that incorporates different sub-graph information, performs poorly as compared to HNZSLP.

\subsection{Impact of Node Number}\label{Impact of Node Number}
\begin{table}[ht]
	\centering
	\renewcommand\arraystretch{1.3}
	\scalebox{0.67}{
	\begin{tabular} {c|c|c|c|c}
\hline
		Nell-ZS No.& $n$&$l$ &$T$ (min)&L\\ 
		\hline
	
		1&13 &30&18&0.1 \\
		2&13&70&37&0.5\\
		3&13&90&45&0.63\\
		4&13&110&66&0.74\\
		\hline
			Wiki-ZS No.& $n$& $l$ &$T$ (min)&L\\ 
		1&15 &30 & 32 &0.15 \\
			2&15 &70 & 46 &0.63 \\
			3&15 &90 & 48 &0.78 \\
			4&15 &110 & 52 &0.89 \\

		\hline
	\end{tabular}}
	\caption{Different node number setting, L rate denotes that there are $\%L$  relations that can cover the whole nodes in the n-gram graph. $T$ refers to the training time. }
\label{con: para_setting}
\end{table}

In our proposed model, there are two kinds of parameters controlling the size of the n-gram graph. one is the gram number $n$, and the other is the node number $l$ in the n-gram graph. In HNZSLP, we treat these two parameters with the same importance.
For the $n$ parameter, we use the maximum word length about the seen relation set as the gram number. In NELL-ZS, $n=13$, in Wiki-ZS, $n=15$. If the word length is smaller than the maximum word length,  the gram number is its word length. 

The performance of our models differs in terms of accuracy and training time under the nodes with different numbers. To investigate the influence of different node sizes, we conduct experiments using HNZSLP with different parameter settings which are shown in  Table \ref{con: para_setting}. We set the batch size to 32 both in NELL-ZS and Wiki-ZS. The training epoch in NELL-ZS is 80, and the training epoch in Wiki-ZS is 70. Other configurations are the same for HNZSLP with different node number settings. We list the training time of each running on GPU. The GPU computations were run on a single Nvidia TITAN RTX. In this section, we experiment with the models under the node number up to 110 for a relation, the KGE model is TransE.
\begin{figure}[t]
        \centering
        \includegraphics[width=0.4\columnwidth]{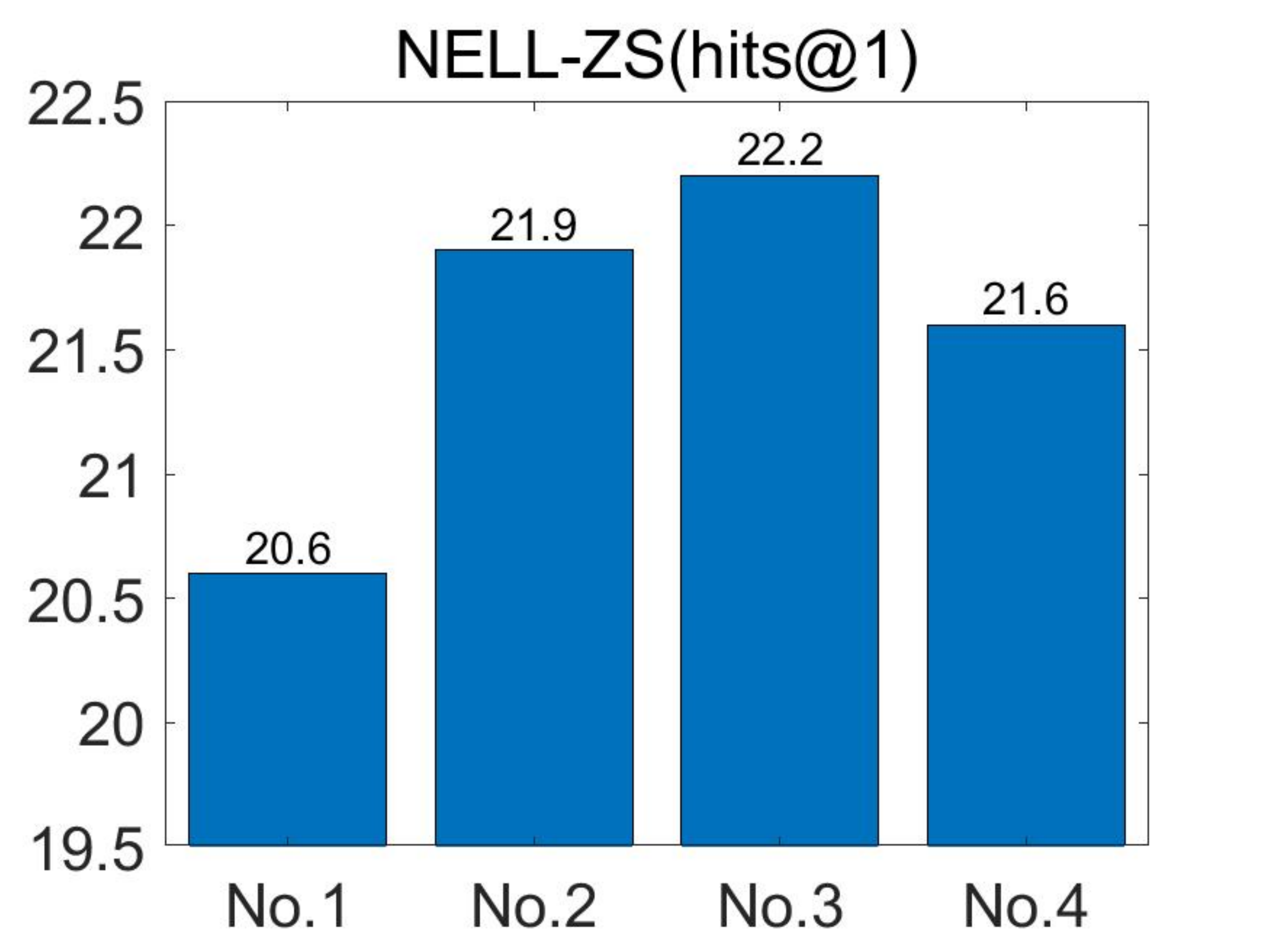}
        \hspace{0.1in}
        \includegraphics[width=0.4\columnwidth]{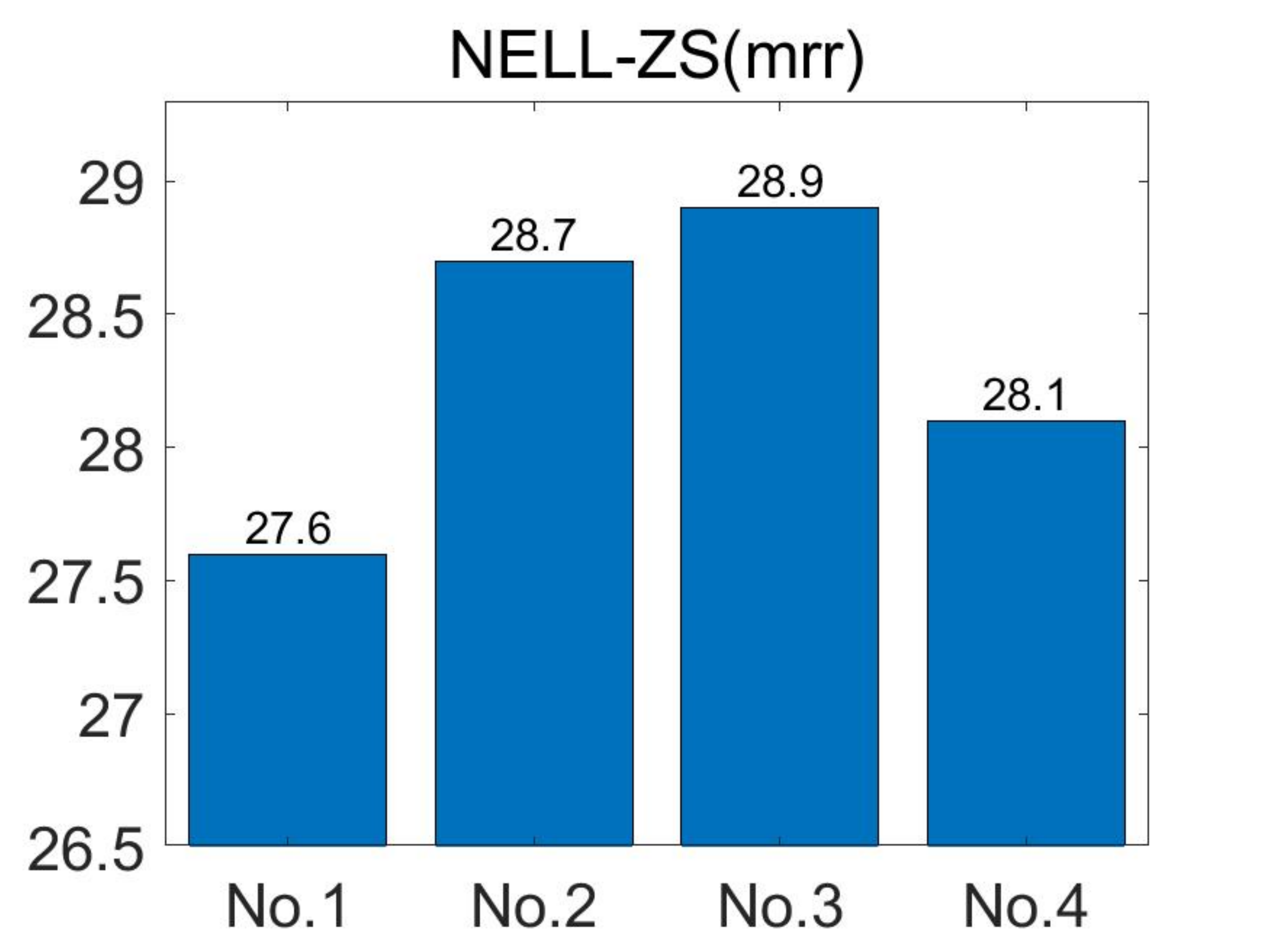}
          \hspace{0.1in}
        \quad
         \includegraphics[width=0.4\columnwidth]{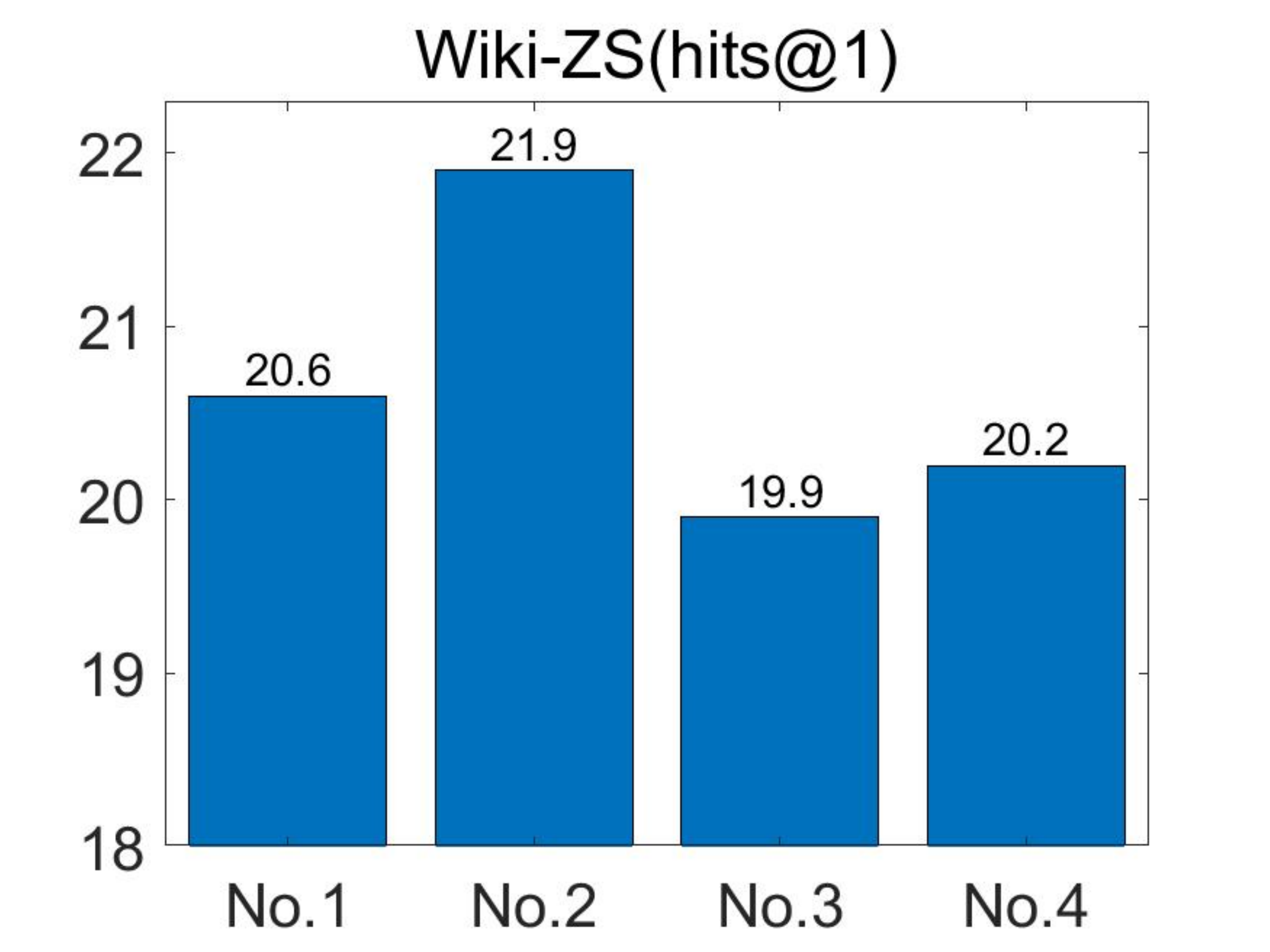}
  \hspace{0.1in}
         \includegraphics[width=0.4\columnwidth]{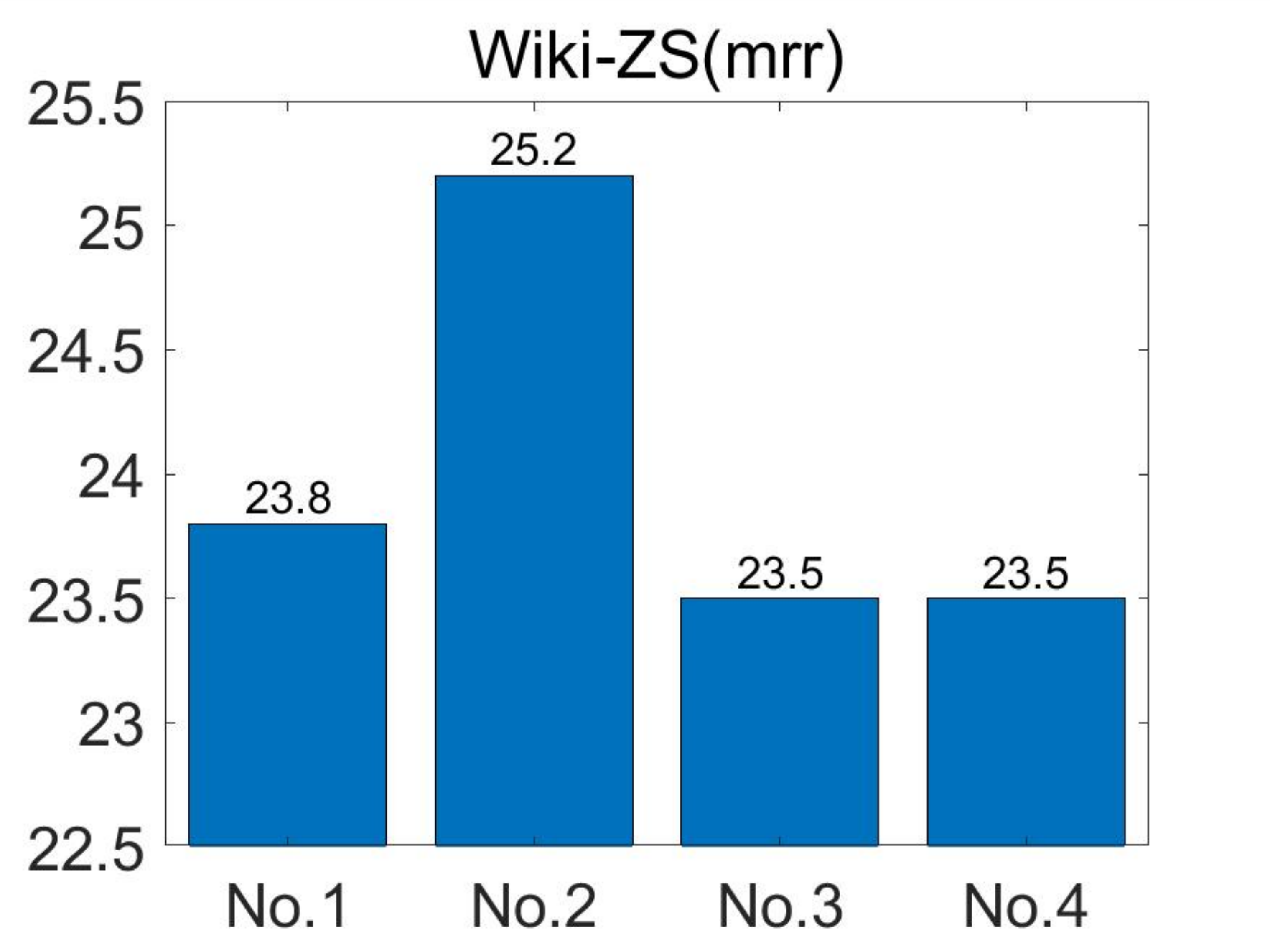}
	\caption{Mrr and hits@1 results on the dataset NELL-ZS and Wiki-ZS with different node numbers}
	\label{con: node numbers}
\end{figure}

From Figure~\ref{con: node numbers}, we can find that, under the fixed setting of gram number $n$, HNZSLP with more nodes achieves higher accuracy, for example, in the dataset NELL-ZS, the hits@1 results with node number 90 are higher than the hits@1 results with node number 30 and 70. This situation indicates that extending the node number is effective.
%%%%%%%%%%%%
However, the performance of our model about No.4 (NELL-ZS, $n$=13, $l$=110) cannot reach the one of No.3  (NELL-ZS,$n$=13, $l$=90) and No.2 (NELL-ZS,$n$=13, $l$=70), despite they have the same gram length $13$.  Our experimental experience suggests that it is not necessary to utilize the whole n-gram graph in HNZSLP, let alone that using the whole n-gram graph may significantly increase the computational cost.
% In our work more node numbers mean lattice graph contains the more coarse-grained node.
In practice, we should choose a suitable number of nodes to build the n-gram graph. However, it is difficult to set appropriate node numbers because there are no systematic methods. We thus set the node numbers by experimental experience. In future work, we will design a more efficient approach to solve this problem and balance the trade-off between the hits@1, hits@5, hits@10, mrr results, and efficiency.

\subsection{Comparison with Different Language Models} \label{Comparison with Different Language Model}

To quantitatively evaluate the effect of  HNZSLP, we compare the performance of  HNZSLP against two models which are based on the latest work about the language model in the zero-shot link prediction task, 
% these works are fine-tuned in the dev dataset, 
and all based on the $BERT_{base}$ model \footnote{https://huggingface.co/bert-base-uncased}. These experiments are conducted with the test set of NELL-ZS. They are described below:
\begin{itemize}
    \item \textbf{KGE-BERT.} We use BERT \cite{devlin2018bert} to calculate the relation embedding $S$ in eq.\ref{con:score}. The KGE score function $f(.)$ (TransE) is utilized to predict the tail entity. We refer to this model as KGE-BERT.
    
    \item \textbf{STAR.} STAR \cite{wang2021structure} is the first work to explore the ability of language models in zero-shot link prediction by using the textural information of entity and relation. In this work, the authors use structured knowledge information and textual information of entity and relation to infer the tail entity. Moreover, they develop a self-adaptive ensemble scheme to improve the model performance by incorporating the triple scores. 
\end{itemize}

\begin{figure}[t]
        \centering
        \includegraphics[width=0.8\columnwidth]{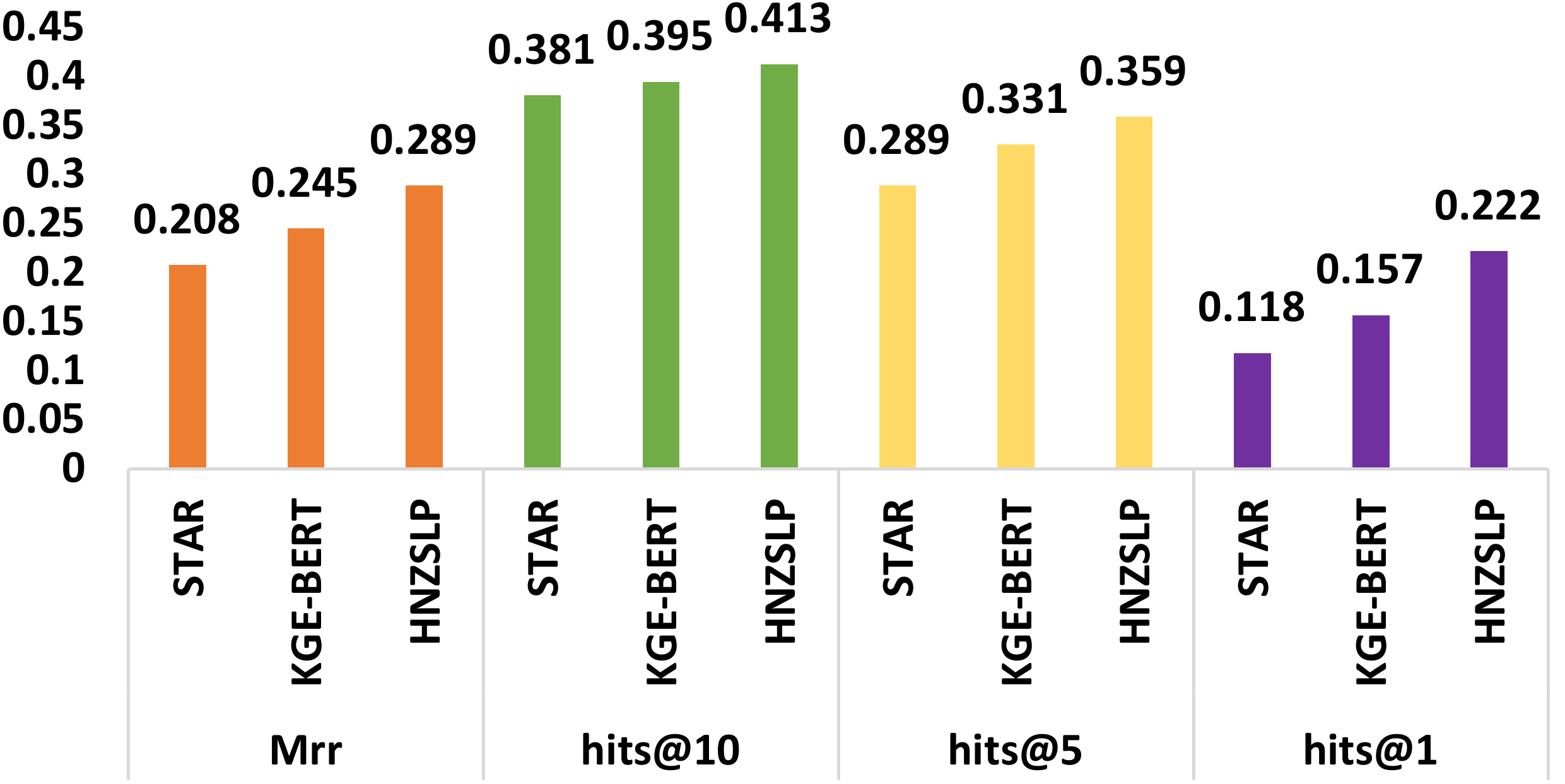}
	\caption{The model performance with different zero shot link prediction methods based on the language model.}
	\label{con:language_model}
\end{figure}

We perform a detailed comparative study on different zero-shot language models to examine their impact on knowledge transfer from seen relation to unseen relation under the KGE framework. Figure~\ref{con:language_model} presents the results with a comparison to KGE-BERT and STAR. 

Figure~\ref{con:language_model} shows that our model can achieve a hits@1 score of 0.214, outperforming the model KGE-BERT which uses the language model to learn the relation information by a large margin of 0.025 hits@1 scores.  For the latest work which uses the language model to infer the tail entity, our model can outperform STAR by a margin of 0.096 hits@1 scores. The advantages of HNZSLP are also verified by metrics MRR, hits@10, and hits@5.

In KGE-BERT, we use the language model to learn the relational textual information, but this way ignores the importance of n-gram graph information. In the model STAR, the authors use the language model to enhance the inference ability of link prediction, they combine the textual information of the head entity and relation by a special token \textit{[SEP]} and then build the triple score with learned tail entity information by the language model. Unfortunately, this  model cannot be used to solve the out-of-vocabulary problem for the current word, though the language model can use its previous knowledge. More importantly, the n-gram graph information is also an important source to improve the performance of tail entity inference.

%% file: 7_conclution.tex
\section{Conclusion}

In this paper, we propose a novel ZSL framework HNZSLP for link prediction. Specifically, we proposed a GramTransformer to learn the n-gram graph information of the relation surface name and utilize the KGE model to infer the tail entity. 
% We evaluate the proposed zero-shot learning framework on two different zero-shot learning tasks.
Experimental results show that our framework achieves consistent improvements over various baselines in two ZSLP datasets. As the GramTransformer can be considered as a text representation method, in the future, we intend to explore its effectiveness on other NLP tasks including text classification.

\section{Limitations}

When the surface name of relations is too long, it means the scale of our build n-gram graph  is large, this way will influence the  efficiency of graph computation. In our work, we just proposed two strategies to select the fixed number of nodes, this way ignores the semantic information about the nodes which are not selected. So in the future, we will design a novel strategy to dynamically select the node, and consider the computational problem at the same time.